\documentclass[runningheads]{llncs}
\usepackage{graphicx}
\usepackage{tabularx} 
\usepackage{amsmath}
\usepackage{amssymb}
\usepackage{multirow}
\usepackage{algorithm}
\usepackage[noend]{algpseudocode}
\usepackage{xcolor}

\newcommand\done[1]{{\color{black}#1}}

\newcommand\norm[1]{\left\lVert#1\right\rVert}

\begin{document}
\title{Overcoming Data Limitation \\in Medical Visual Question Answering}
\titlerunning{Overcoming Data Limitation in Medical VQA}
%
\author{Binh D. Nguyen\inst{1}\and
Thanh-Toan Do\inst{2}\and
Binh X. Nguyen\inst{1}\and
Tuong Do\inst{1}\and\\
Erman Tjiputra\inst{1}\and
Quang D. Tran\inst{1}}

%
\institute{AIOZ Pte Ltd, Singapore \\ \email{\{binh.duc.nguyen,binh.xuan.nguyen,tuong.khanh-long.do,\\erman.tjiputra,quang.tran\}@aioz.io} \and University of Liverpool \\
\email{thanh-toan.do@liverpool.ac.uk}}

\maketitle              
%
\begin{abstract}
\done{
Traditional approaches for Visual Question Answering (VQA) require large amount of labeled data for training. Unfortunately, such large scale data is usually not available for medical domain. In this paper, we propose a novel medical VQA framework that overcomes the labeled data limitation. The proposed framework explores the use of the unsupervised Denoising Auto-Encoder (DAE) and the supervised Meta-Learning. The advantage of DAE is to leverage the large amount of unlabeled images while the advantage of Meta-Learning is to learn meta-weights that quickly adapt to VQA problem with limited labeled data. By leveraging the advantages of these techniques, it allows the proposed framework to be efficiently trained using a small labeled training set. The experimental results show that the proposed method significantly outperforms the state-of-the-art medical VQA. The source code is available at \url{https://github.com/aioz-ai/MICCAI19-MedVQA}.
}
\keywords{Visual Question Answering\and Auto-Encoder\and Meta-Learning.}
\end{abstract}
\section{Introduction}
\label{Sec:Intro}
\done{
Visual Question Answering (VQA) aims to provide a correct answer to a given question such that the answer is consistent with the visual content of a given image. In medical domain, VQA could benefit both doctors and patients. For example, doctors could use answers provided by VQA system as support materials in decision making, while patients could ask VQA questions related to their medical images for better understanding their health. However, one major problem with medical VQA is the lack of large scale labeled training data which usually requires huge efforts to build. The first attempt for building the dataset for medical VQA is by ImageCLEF-Med~\cite{ImageCLEFVQAMed2018}. In particular, in~\cite{ImageCLEFVQAMed2018}, images were automatically captured from PubMed Central articles. The questions and answers were automatically generated from corresponding captions of images. By that construction, the data has high noisy level, i.e., the dataset includes many images that are not useful for direct patient care and it also contains questions that do not make any sense. Recently, in~\cite{lau2018dataset}, the authors released the first manually constructed dataset VQA-RAD for medical VQA. Unfortunately, it contains only 315 images, which prevents to directly apply the powerful deep learning models for the VQA problem. One may think about the use of transfer learning  in which the pretrained deep learning models~\cite{Simonyan2015VGG,He2016ResNet} that are trained on the large scale labeled dataset such as ImageNet~\cite{russakovsky2015imagenet} are used for finetuning on the medical VQA. However, due to difference in visual concepts between ImageNet images and medical images, finetuning with very few medical images is not sufficient, which is confirmed by our experiments in Section~\ref{Sec:Exp}. Therefore it is necessary to develop a new VQA framework that can improve the accuracy while still only needs a small labeled training data. 
}

\done{
The motivation for our approach to overcome the data limitation of medical VQA comes from two observations.
Firstly, we observe that there are large scale unlabeled medical images available. These images are from same domain with medical VQA images.  Hence if we train an unsupervised deep learning model using these unlabeled images, the trained weights may be easier to be adapted to the medical VQA problem than the pretrained weights on ImageNet images. Another observation is that although the labeled dataset VQA-RAD in~\cite{lau2018dataset} is primarily designed for VQA, by spending a little effort, we can extract the new class labels\footnote{The descriptions of new defined classes are presented in Section~\ref{Sec:Dataset}.} for that dataset. The new class labels allow us to apply the recent meta-learning technique \cite{finn2017model} for learning meta-weights, that can be quickly adapted to the VQA problem later. 

From these two observations, we propose a novel medical VQA framework as presented in Figure~\ref{fig:framework}, in which the Model-Agnostic Meta-Learning (MAML) \cite{finn2017model} and the Convolutional Denoising Auto-Encoder (CDAE) \cite{masci2011stacked} are used to initialize the model weights for the image feature extraction. 
}
\section{Literature Review}
\label{Sec:Literature}
\done{
\textbf{Medical Visual Question Answering.}
Most approaches for medical VQA~\cite{lau2018dataset,Peng2018UMass,Abacha2018NML,zhou2018InceptionResnet} are to directly apply the state-of-the-art general VQA models to medical domain. The 2018 ImageCLEF-Med challenge~\cite{ImageCLEFVQAMed2018} provides a good overview about the approaches and their results.  \done{Typically, in~\cite{Peng2018UMass,Abacha2018NML,zhou2018InceptionResnet}, the authors use the state-of-the-art attention mechanisms in general VQA (e.g., MCB \cite{fukui2016multimodal}, SAN \cite{Yang2016StackedAN}) to learn a join representation between an image and a question. 
Note that in the mentioned approaches, the models pretrained on ImageNet such as VGG \cite{Simonyan2015VGG} or ResNet \cite{He2016ResNet} are directly finetuned on medical VQA images for image feature extraction. However, directly  finetuning those models on  medical VQA images is not effective due to the limited medical VQA data. 
}
}
\\
\done{
\textbf{Meta-learning.}  
Traditional machine learning algorithms, especially deep learning based approaches, require large scale labeled training set when learning a new task, even when the model is pretrained on other classification  problems~\cite{maicas2018training,DBLP:conf/micad/BarDWG15}. Contrasting with traditional machine learning algorithms, meta-learning approach~\cite{schmidhuber1987evolutionary} targets to deal with the problem of data limitation when  learning new tasks. Recently, in~\cite{finn2017model} the authors proposed a new approach for meta-learning, i.e.,  Model-Agnostic Meta-Learning (MAML), which helps to learn a meta-model (e.g. network weights) from current tasks that is broadly suitable for many tasks. Hence, the model can be quickly adapted  to new tasks that have a small number of training images.
}
\\
\done{
\textbf{Denoising Auto-Encoder.}
In medical domain, the lack of labeled data makes training process become inefficiency. Thus, unlabeled data, which is easy to achieve, is encouraged to use for training. 
Auto-Encoder \cite{rumelhart1985learning,masci2011stacked}, which helps to extract high-level features without any label information, is a typical solution to take the advantage of unlabeled data.}
\done{Besides, medical images such as MRI, CT, X-ray may contain various degree of noises, which might happen during transmission and acquisition \cite{jifara2017medical}. Hence, it requires a feature extraction model that is robust to noise, i.e., it can still extract useful information from the noisy input image. In this work, to leverage the benefit of large scale unlabeled datasets and also to make the model robust to the noise in input images, we propose to use the Convolutional Denoising Auto-Encoder (CDAE)~\cite{masci2011stacked} as one of image feature extraction components in our framework.
}
\section{Methodology}
\label{Sec:Method}
 \done{
 The proposed medical VQA framework is presented in Figure~\ref{fig:framework}. In our framework, the image feature extraction component is initialized by pretrained weights from MAML and CDAE. After that, the VQA framework will be finetuned in an end-to-end manner on the medical VQA data. In the following sections, we detail the architectures of MAML, CDAE, and our framework.
 }
\done{
\subsection{Model-Agnostic Meta-Learning -- MAML}
\label{subsec:MAML}

\begin{algorithm}[!t]
\caption{Overview of the meta-training procedure}\label{alg:maml}
\begin{algorithmic}[1]
\Procedure{META-TRAIN}{$\mathcal{D}$, model $f_\theta$}
\State Initialize model parameters $\theta$
\For{$h=1$ $to$ $H$}\Comment{Meta-update Loop}
\State \textbf{Create} meta-batch of tasks $\{\mathcal{T}_1, \mathcal{T}_2, ..., \mathcal{T}_m\}$
\For{each task $\mathcal{T}_i$}
\State Sample data \{$\mathcal{D}^{tr}_i, \mathcal{D}^{val}_i$\} of task $\mathcal{T}_i$
\State Update task models with Eq. (\ref{eq:1}) using samples from $\mathcal{D}^{tr}_i$ 
\EndFor
\State \textbf{Update} meta-model $\theta$ with Eq. (\ref{eq:2}) using $\{\mathcal{D}^{val}_1, \mathcal{D}^{val}_2, ..., \mathcal{D}^{val}_m\}$
\EndFor
\EndProcedure
\end{algorithmic}
\end{algorithm}

The MAML classification model is represented   by   a parametrized function $f_{\theta}$ with meta-parameters $\theta$. When adapting to a new task $\mathcal{T}_i$, the model's parameters $\theta$ become $\theta'_i$.
Let  $\mathcal{D} = \{x_i, y_i\}_{i=1}^{N}$ be the dataset for training MAML. $N$ is the number of samples.
$\{x_i,y_i\}$ is a pair of image ($x_i$) and its class label ($y_i$). 
A task in MAML is defined as a ``\textbf{k}-shot \textbf{n}-way'' classification problem. 
The dataset for each task is defined as $\mathcal{D}' = \{x'_i, y'_i\}_{i=1}^{N'}$; samples in $\mathcal{D}'$ come from $n$ different classes which are a subset of classes in $\mathcal{D}$. 
The task dataset $\mathcal{D}'$ is split equally into two sets $\mathcal{D}^{tr}$ -- training set and $\mathcal{D}^{val}$ -- validation set;  in  $\mathcal{D}^{tr}$, each class contains $k$ training images.  
The training procedure is  described in the Algorithm \ref{alg:maml}. In each iteration $h$, $m$ tasks are generated forming a meta-batch for MAML training. For each task $\mathcal{T}_i$, the corresponding adapted parameters $\theta_i'$ are calculated as follows
\newcommand{\Lagr}{\mathcal{L}}
\begin{equation}
    \theta'_i = \theta - \alpha\nabla_\theta L_{\mathcal{T}_i}(f_\theta (\mathcal{D}^{tr}_i))\label{eq:1}
\end{equation}
where $L_{\mathcal{T}_i}$ is the classification loss of task $i$. After all adapted parameters of $m$ tasks are calculated, the meta-model's parameters $\theta$ are updated via \textit{stochastic gradient descent} (SGD) as follows
\begin{equation}
    \theta \gets \theta - \beta\nabla_\theta\sum_{\mathcal{T}_i} L_{\mathcal{T}_i}(f_{\theta'_{i}}(\mathcal{D}^{val}_i)) \label{eq:2}
\end{equation}
We follow~\cite{finn2017model} to design MAML. It consists of four $3\times3$ convolutional layers with stride $2$ and is ended with a mean pooling layer; each convolutional layer has $64$ filters and is followed by a ReLu layer. 
The detail training of MAML is presented in Section~\ref{Sec:ExpDetail}. After training, the weights of the meta-model are used  for finetuning in the VQA framework as presented in Figure~\ref{fig:framework}.
}
\done{
\subsection{Convolutional Denoising Auto-Encoder -- CDAE}
\label{subsec:CDAE}
The encoder maps an image $x'$, which is the noisy version of the original image $x$, to a latent representation $z$ which retains  useful amount of information. The decoder transforms $z$ to the output $y$. The training algorithm aims to minimize the reconstruction error between $y$ and the original image $x$ as follows
\begin{equation}
L_{rec} = \norm{x-y}_2^2
 \label{eq:rec_loss}
\end{equation}
In our design, the encoder is a stack of convolutional layers; each of them is followed by a max pooling layer. The decoder is a stack of  deconvolutional and convolutional layers. The noisy version $x'$ is achieved by adding Gaussian noise to the original image $x$. 
The detail training of CDAE is presented in Section~\ref{Sec:ExpDetail}. After training, the trained weights of both encoder and decoder are used for finetuning in the VQA framework as presented in Figure~\ref{fig:framework}.
\subsection{The proposed Medical VQA framework}
\begin{figure*}
    \centering
    \includegraphics[width=\textwidth, keepaspectratio=true]{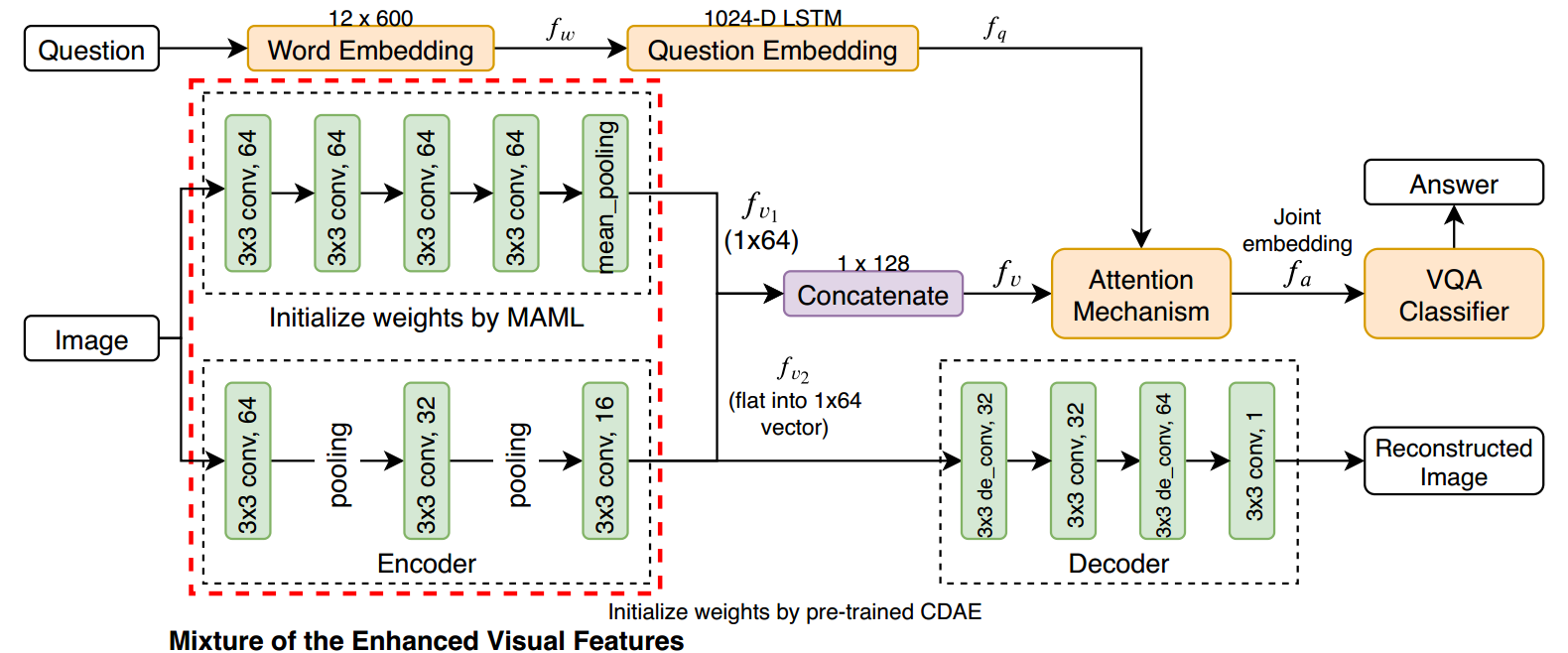}
    \caption{The proposed medical VQA. The  image feature extraction is denoted as ``Mixture of Enhanced Visual Features (MEVF)'' and is marked with the red dashed box. The weights of MEVF are intialized by MAML and CDAE. Best view in color.}
    \label{fig:framework}
\end{figure*}
}
\done{
\textbf{VQA detail.}
Each input question is trimmed to a 12-word sentence. The question is zero-padded in case its length is less than 12. Each word is represented by a 600-D vector which is a concatenation of the 300-D GloVe word embedding \cite{pennington2014glove} 
and the augmenting embedding from the VQA-RAD training data \cite{Kim2018BilinearAN}. The word embedding is fed into a 1024-D LSTM in order to produce the question embedding, denoted as $f_q$ in Figure~\ref{fig:framework}. Each input image is passed through the Mixture of Enhanced Visual Features (MEVF) component, which produces two 64-D vectors $f_{v1}$ and $f_{v2}$. Those vectors are concatenated to form an $128$-D enhanced image feature, denoted as $f_v$ in Figure~\ref{fig:framework}.

Image feature $f_v$ and question embedding $f_q$ are fed into an attention mechanism (BAN~\cite{Kim2018BilinearAN} or SAN~\cite{Yang2016StackedAN}) to produce a joint representation  $f_a$. This feature $f_a$ is used as input for a multi-class classifier (over the set of predefined answer classes~\cite{lau2018dataset}). To train the proposed model, we introduce a multi-task loss function to incorporate the effectiveness of the CDAE to VQA. Formally, our loss function is defined as follows
\begin{equation}
    L = \alpha_1 L_{vqa} + \alpha_2 L_{rec}
\end{equation}
where $L_{vqa}$ is a Cross Entropy loss for VQA classification and $L_{rec}$ stands for the reconstruction loss of CDAE  (Eq. \ref{eq:rec_loss}). The whole VQA model is finetuned in an end-to-end manner using VQA-RAD dataset as presented in Section~\ref{Sec:ExpDetail}.
}

\section{Experiments}
\label{Sec:Exp}
\done{
\subsection{Dataset}
\label{Sec:Dataset}
The VQA-RAD~\cite{lau2018dataset} dataset contains 315 images and 3,515 corresponding questions. Each image is associated with more than one question. The questions are divided into {11} categories which are ``Abnormality'', ``Attribute", ``Color'', ``Count'', ``Modality'', ``Organ'', ``Other'', ``Plane'', ``Positional reasoning'', ``Object/Condition Presence'', ``Size''. We use exactly the same training set and test set described in \cite{lau2018dataset}. The test set contains 451 questions and the rest is for training. The questions can be close-ended questions, i.e. the questions in which the answers are  ``yes/no'' and other limited choices, or open-ended questions, i.e., the questions do not have a limited structure and could have multiple correct answers. The dataset has 458 answers. The VQA is posed as a classification over the set of answers. 
}
\subsection{Training MAML, CDAE, and the whole VQA framework}
\label{Sec:ExpDetail}
\done{
\textbf{MAML.} We create the  dataset for training MAML by \textit{manually reviewing} around three thousand question-answer pairs from the training set of VQA-RAD dataset. In our annotation process, images are split into three parts based on its \textit{body part} labels (head, chest, abdomen). Images from each body part are further divided into three subcategories based on the interpretation from the question-answer pairs corresponding to the images. These subcategories are: (1) \textit{normal} images in which no pathology is found; (2) \textit{abnormal present} images in which there are the existence of fluid, air, mass, or tumor; (3) \textit{abnormal organ} images in which the organs are large in size or in wrong position}. Thus, all the images are categorized into 9 classes: \textit{head normal, head abnormal present, head abnormal organ, chest normal, chest abnormal organ, chest abnormal present, abdominal normal, abdominal abnormal organ, and abdominal abnormal present}. For every iteration of MAML training (line 3 in Alg.~\ref{alg:maml}),  5 tasks are sampled per iteration. For each task, we randomly select 3 classes (from 9 classes). For each class, we randomly select 6 images in which $3$ images are used for updating task models and the remaining $3$ images are used for updating meta-model.
\\
\done{
\textbf{CDAE.} 
To train CDAE, we collect $11,779$ unlabeled images available online
which are brain MRI images  \cite{clark2013cancer}, chest X-ray images\footnote{https://www.kaggle.com/paultimothymooney/chest-xray-pneumonia} and CT abdominal images \footnote{https://www.synapse.org/\#!Synapse:syn3193805/wiki/217753}. 
The dataset is split into train set with $9,423$ images and test set with $2,356$ images. We use Gaussian noise to corrupt the input images before feeding them to the encoder.
}
\\
\done{
\textbf{VQA.} 
After training MAML and CDAE, we use their trained weights to initialize the MEVF image feature extraction component in the VQA framework. We then finetune the whole VQA model using the training set of VQA-RAD dataset. In order to make a fair comparison to \cite{lau2018dataset}, we evaluate our framework on $300$ free-form questions of the test set. The  proposed framework is implemented using PyTorch. The experiments are conducted on a single NVIDIA 1080Ti with 11GB RAM. The VQA accuracy is computed as the percentage of the total correct answers over the number of testing questions.}
\subsection{Ablation Study}
\done{We evaluate the effectiveness of different image feature extraction methods in the VQA model when using only MAML, using only CDAE, and their combination MEVF. For each extraction method, we present results when training the VQA model using only  VQA-RAD training set (i.e. \textit{from scratch}) or when pretraining as described in Section~\ref{Sec:ExpDetail} and then finetuning using VQA-RAD training set (i.e. \textit{finetuning}). We also present the results when the pretrained VGG model (on ImageNet) is finetuned on VQA-RAD for image feature extraction.}

\done{
Table \ref{tab:ablation_study} presents VQA accuracy in both VQA-RAD open-ended and close-ended questions on the test set. The results  show that for both MAML and CDAE, by firstly pretraining as described in Section~\ref{Sec:ExpDetail}, then finetuning, the finetuning significantly improves the performance over the training from scratch using only VQA-RAD. 
In addition, the results also show that our pretraining and finetuning of MAML and CDAE give better performance than the finetuning of VGG-16 which is pretrained on the  ImageNet dataset. Our proposed image feature extraction MEVF which leverages both pretrained weights of MAML and CDAE, then finetuning them give the best performance. This confirms the effectiveness of the proposed MEVF for  dealing with the limitation of labeled training data for medical VQA. 
}
\begin{table}[!t]
\centering
\small
\caption{VQA results on VQA-RAD test set. All reference methods differ at the image feature extraction component. Other components are similar. The Stacked Attention Network (SAN) \cite{Yang2016StackedAN} is used as the attention mechanism in all methods.}
\begin{tabular}{|l|c|c|}
\hline
\multicolumn{1}{|c|}{\multirow{2}{*}{\textbf{Reference  methods}}} & \multicolumn{2}{c|}{\textbf{VQA accuracy (\%)}} \\ \cline{2-3} 
\multicolumn{1}{|c|}{}                                             & \textbf{Open-ended}  & \textbf{Close-ended}  \\ \hline
VGG-16 (finetuning)\cite{lau2018dataset}          & 24.2                   & 57.2                   \\ \hline
MAML (from scratch)                                                & 6.5                    & 68.6                   \\ 
\textbf{MAML(finetuning)}                                          & \textbf{38.2}          & \textbf{69.7}          \\ \hline
CDAE (from scratch)                                                & 13.8                   & 69.2                   \\ 
\textbf{CDAE (finetuning)}                                         & \textbf{36.7}          & \textbf{70.8}          \\ \hline
MEVF (from scratch)                                                & 15.4                   & 70.8                   \\ 
\textbf{MEVF (finetuning)}                                         & \textbf{40.7}          & \textbf{74.1}          \\ \hline
\end{tabular}
\label{tab:ablation_study}
\end{table}

\done{
The results also show that for all MAML, CDAE, and MEVF, the accuracy on close-ended questions (CEQ) are  higher than those on the open-ended questions (OEQ).  Furthermore, the improvements of the finetuning over the training from scratch are more significant on OEQ. We found that OEQ are usually difficult to answer than CEQ, i.e., OEQ mainly ask about the detail description and require long answers, while CEQ mainly ask about the confirmation (i.e., ``yes/no'') and usually have short answers. 
That observation implies that the description answers which need more information from input images take more benefits from the proposed image feature extraction.
}
\begin{table}[!t]
\centering
\footnotesize
\caption{Performance comparison on VQA-RAD test set. The results of SAN framework (fw.) and MCB framework (fw.) are cited from the paper~\cite{lau2018dataset}.}
\begin{tabular}{|c|c|c|c|c|c|}
\hline
                                       &   \textbf{\begin{tabular}[c]{@{}c@{}}SAN fw.\cite{lau2018dataset,Yang2016StackedAN}\\  (baseline)\end{tabular}} & \textbf{\begin{tabular}[c]{@{}c@{}}MCB fw.\cite{lau2018dataset,fukui2016multimodal}\\ (baseline)\end{tabular}} & \textbf{\begin{tabular}[c]{@{}c@{}}BAN fw.\cite{Kim2018BilinearAN}\\ (baseline)\end{tabular}} & \textbf{\begin{tabular}[c]{@{}c@{}}SAN +\\ proposal\end{tabular}} & \textbf{\begin{tabular}[c]{@{}c@{}}BAN +\\ proposal\end{tabular}} \\ \hline

Open-ended & 24.2    & 25.4  & 27.6  & 40.7 &\textbf{43.9}  \\ \hline
Close-ended &  57.2    & 60.6    & 66.5   & 74.1 & \textbf{75.1} \\ \hline

\end{tabular}
\label{tab:sota_open}
\end{table}
\subsection{Comparison with the state of the art}
\done{
We compare our framework (Figure~\ref{fig:framework}) with the baselines in~\cite{lau2018dataset}. In~\cite{lau2018dataset}, the authors report the results when applying the general VQA frameworks, i.e., SAN framework~\cite{Yang2016StackedAN}, MCB framework~\cite{fukui2016multimodal}\footnote{Those frameworks are completed VQA models in which the core components in those frameworks are SAN and MCB attentions. We refer the reader to the corresponding papers~\cite{Yang2016StackedAN,fukui2016multimodal} for the detail of those models.} and finetuning on VQA-RAD dataset. We also report another strong baseline when finetuning the state-of-the-art BAN framework~\cite{Kim2018BilinearAN}. 
For our framework, we report results when using SAN~\cite{Yang2016StackedAN} or BAN~\cite{Kim2018BilinearAN} as the attention mechanisms, although other attention mechanisms are straightforward to use in our framework.}

\done{
Table~\ref{tab:sota_open} presents comparative results between methods. Note that for the image feature extraction, the baselines use the pretrained models (VGG or ResNet) that have been trained on ImageNet and then finetune on the VQA-RAD dataset. For the question feature extraction, all baselines and our framework use the same pretrained models (i.e., Glove~\cite{pennington2014glove}) and finetuning on VQA-RAD. The results show that when BAN or SAN is used as the attention mechanism in our framework, it significantly outperforms the baseline frameworks  BAN~\cite{Kim2018BilinearAN} and SAN~\cite{lau2018dataset,Yang2016StackedAN}. Our best setting, i.e. the one with BAN as the attention, achieves the state-of-the-art results and it significantly outperforms the best baseline framework  BAN~\cite{Kim2018BilinearAN}, i.e., the improvements are $16.3\%$ and $8.6\%$ on open-ended and close-ended VQA, respectively.
}

\section{Conclusion}
\label{Sec:Conclusion}
In this paper, we proposed a novel medical VQA framework that leverages the meta-learning MAML and denoising auto-encoder CDAE for image feature extraction in order to overcome the limitation of labeled training data.  
Specifically, CDAE helps to leverage information from the large scale unlabeled images, while MAML helps to learn meta-weights that can be quickly adapted to the VQA problem. We establish new state-of-the-art results on VQA-RAD dataset for both close-ended and open-ended questions.

\bibliographystyle{splncs04} 
\bibliography{paper198.bbl} 

\begin{thebibliography}{10}
\providecommand{\url}[1]{\texttt{#1}}
\providecommand{\urlprefix}{URL }
\providecommand{\doi}[1]{https://doi.org/#1}

\bibitem{Abacha2018NML}
Abacha, A.B., Gayen, S., Lau, J.J., Rajaraman, S., Demner-Fushman, D.: {NLM} at
  {ImageCLEF} 2018 visual question answering in the medical domain. {CEUR}
  Workshop Proceedings (2018)

\bibitem{DBLP:conf/micad/BarDWG15}
Bar, Y., Diamant, I., Wolf, L., Greenspan, H.: Deep learning with non-medical
  training used for chest pathology identification. In: Medical Imaging:
  Computer-Aided Diagnosis (2015)

\bibitem{clark2013cancer}
Clark, K., Vendt, B., Smith, K., Freymann, J., et.al.: {The Cancer Imaging
  Archive (TCIA)}: maintaining and operating a public information repository.
  Journal of Digital Imaging pp. 1045--1057 (2013)

\bibitem{finn2017model}
Finn, C., Abbeel, P., Levine, S.: Model-agnostic meta-learning for fast
  adaptation of deep networks. In: ICML (2017)

\bibitem{fukui2016multimodal}
Fukui, A., Park, D.H., Yang, D., Rohrbach, A., Darrell, T., Rohrbach, M.:
  Multimodal compact bilinear pooling for visual question answering and visual
  grounding. In: EMNLP (2016)

\bibitem{ImageCLEFVQAMed2018}
Hasan, S.A., Ling, Y., Farri, O., Liu, J., Lungren, M., M{\"u}ller, H.:
  Overview of the {ImageCLEF} 2018 medical domain visual question answering
  task. {CEUR} Workshop Proceedings (2018)

\bibitem{He2016ResNet}
He, K., Zhang, X., Ren, S., Sun, J.: Deep residual learning for image
  recognition. In: CVPR (2016)

\bibitem{jifara2017medical}
Jifara, W., Jiang, F., Rho, S., Cheng, M., Liu, S.: Medical image denoising
  using convolutional neural network: a residual learning approach. The Journal
  of Supercomputing pp. 1--15 (2017)

\bibitem{Kim2018BilinearAN}
Kim, J.H., Jun, J., Zhang, B.T.: Bilinear attention networks. In: NIPS (2018)

\bibitem{lau2018dataset}
Lau, J.J., Gayen, S., Abacha, A.B., Demner-Fushman, D.: A dataset of clinically
  generated visual questions and answers about radiology images. Nature  (2018)

\bibitem{maicas2018training}
Maicas, G., Bradley, A.P., Nascimento, J.C., Reid, I., Carneiro, G.: Training
  medical image analysis systems like radiologists. In: MICCAI (2018)

\bibitem{masci2011stacked}
Masci, J., Meier, U., Cire{\c{s}}an, D., Schmidhuber, J.: Stacked convolutional
  auto-encoders for hierarchical feature extraction. In: ICANN (2011)

\bibitem{Peng2018UMass}
Peng, Y., Liu, F., Rosen, M.P.: Umass at imageclef medical visual question
  answering (med-vqa) 2018 task. {CEUR} Workshop Proceedings (2018)

\bibitem{pennington2014glove}
Pennington, J., Socher, R., Manning, C.D.: Glove: Global vectors for word
  representation. In: EMNLP (2014)

\bibitem{rumelhart1985learning}
Rumelhart, D.E., Hinton, G.E., Williams, R.J.: Learning internal
  representations by error propagation. Tech. rep. (1985)

\bibitem{russakovsky2015imagenet}
Russakovsky, O., Deng, J., Su, H., et.al.: Imagenet large scale visual
  recognition challenge. IJCV pp. 211--252 (2015)

\bibitem{schmidhuber1987evolutionary}
Schmidhuber, J.: Evolutionary Principles in Self-referential Learning. (1987)

\bibitem{Simonyan2015VGG}
Simonyan, K., Zisserman, A.: Very deep convolutional networks for large-scale
  image recognition. In: ICLR (2015)

\bibitem{Yang2016StackedAN}
Yang, Z., He, X., Gao, J., Deng, L., Smola, A.J.: Stacked attention networks
  for image question answering. In: CVPR (2016)

\bibitem{zhou2018InceptionResnet}
Zhou, Y., Kang, X., Ren, F.: Employing {Inception-Resnet-v2} and {Bi-LSTM} for
  medical domain visual question answering. {CEUR} Workshop Proceedings (2018)

\end{thebibliography}
\end{document}